\documentclass[conference,a4paper]{IEEEtran}

\usepackage{amsmath,amsfonts,amssymb}
\usepackage{algorithmic}
\usepackage{makecell}
\usepackage{array}
\usepackage{textcomp}
\usepackage{bbm}

\usepackage{stfloats}
\usepackage{subcaption}
\usepackage[ruled,linesnumbered]{algorithm2e}
\usepackage{url}
\usepackage{verbatim}
\usepackage{graphicx}
\usepackage{cite}
\usepackage{color}
\usepackage{booktabs}
\usepackage{multirow}
\def\BibTeX{{\rm B\kern-.05em{\sc i\kern-.025em b}\kern-.08em
    T\kern-.1667em\lower.7ex\hbox{E}\kern-.125emX}}
\IEEEoverridecommandlockouts
\begin{document}

\title{Multi-Agent Actor-Critic with Harmonic Annealing Pruning for Dynamic Spectrum Access Systems}
% \title{Sparse Multi-Agent Actor-Critic Reinforcement Learning for Dynamic Spectrum Access}

\author{\IEEEauthorblockN{George Stamatelis\IEEEauthorrefmark{1}, Angelos-Nikolaos Kanatas\IEEEauthorrefmark{2}, and George C. Alexandropoulos\IEEEauthorrefmark{1}}
\IEEEauthorblockA{\IEEEauthorrefmark{1}Department of Informatics and Telecommunications, National and Kapodistrian University of Athens,\\
Panepistimiopolis Ilissia, 16122 Athens, Greece\\
\IEEEauthorrefmark{2}School of Electrical and Computer Engineering, National Technical University of Athens,\\
Zografou Campus, 15772 Athens, Greece\\
e-mails: \{georgestamat, alexandg\}@di.uoa.gr, el19169@mail.ntua.gr \\
}

\thanks{This work has been supported by the SNS JU project 6G-DISAC under the EU’s Horizon Europe research and innovation program under Grant Agreement No 101139130.}
}


\maketitle

\begin{abstract}  
Multi-Agent Deep Reinforcement Learning (MADRL) has emerged as a powerful tool for optimizing decentralized decision-making systems in complex settings, such as Dynamic Spectrum Access (DSA). However, deploying deep learning models on resource-constrained edge devices remains challenging due to their high computational cost. To address this challenge, in this paper, we present a novel sparse recurrent MARL framework integrating gradual neural network pruning into the independent actor global critic paradigm. Additionally, we introduce a harmonic annealing sparsity scheduler, which achieves comparable, and in certain cases superior, performance to standard linear and polynomial pruning schedulers at large sparsities. Our experimental investigation demonstrates that the proposed DSA framework can discover superior policies, under diverse training conditions, outperforming conventional DSA, MADRL baselines, and state-of-the-art pruning techniques.
\end{abstract}


\section{Introduction}\label{sec:intro}
Modern wireless communications are witnessing a rapid surge in the number of devices contending for limited spectrum resources. Since the available spectrum is inherently sparse, this increased demand underscores the need for effective spectrum management schemes. In this paper, we focus on the Dynamic Spectrum Access (DSA) problem, specifically, its distributed variant, where multiple users broadcast messages over a shared set of orthogonal channels to maximize network utility, all without direct coordination. Upon transmitting on a given channel, a user receives a binary observation indicating whether the transmission succeeded or failed due to interference from others, as well as information about the transmit Signal-to-Noise Ratio (SNR). Consequently, each user must not only determine the most suitable channels for their location and equipment, but also infer the access strategies of their peers to minimize collisions and optimize channel utilization.

Deep Reinforcement Learning (DRL) algorithms have emerged as a powerful candidate for finding channel allocation policies, but their inherent issues regarding neural network over-parameterization, storage, and computation requirements, especially when being deployed in lightweight Internet of Things (IoT) devices, remain largely unaddressed.

\subsection{Background}\label{subsec:background}
\paragraph{DRL for DSA}
Dynamic spectrum management problems, including DSA, have been thoroughly studied in~\cite{SpecSharing5gSurvey}, and optimal solutions have been developed for certain special cases (e.g., \cite{optimDSA1}). Due to the difficulty of realistic DSA problems, various methods for finding approximately optimal policies have been proposed, including greedy algorithms \cite{zhaoPODMPGreedi}, conventional optimization (i.e., without deep neural networks), reinforcement learning \cite{singleandMarlDSA}, and bandits \cite{banditsDSA2}. Recent research on DSA problems (e.g., \cite{DRL-multichannel}) applies existing DRL algorithms for Partially Observable Markov Decision Processes (POMDPs), with elaborate modifications to handle the unique DSA challenges.
% Recent research on DSA problems applies existing DRL algorithms for Partially Observable Markov Decision Processes (POMDP)s, with elaborate modifications to handle the unique challenges of this problem.
% For instance, \cite{DRL-multichannel} applies a  single-agent DRL algorithm to DSA problems with great success.
Distributed Multi-Agent DRL (MADRL) algorithms, using the Centralized Training with Decentralized Execution (CTDE) principle~\cite{marlRehearsal}, have been also studied in~\cite{DDSA-Cohen,DSAEurasip,DDSAandControl}. The existing approaches for MARL-based DSA fall into three main categories. The first involves \textit{Deep Q-Networks (DQN) with parameter sharing}, where all agents share a single Q-network during training and deploy identical clones \cite{DDSA-Cohen,DNQnature}. The second category includes \textit{Actor-Critic (AC) methods with full parameter sharing}, applied to dynamic channel access and power control \cite{DDSAandControl}. Finally, the \textit{Independent Actor with Global Critic (IAGC) paradigm} allows each agent to maintain an independent actor while sharing a centralized critic,~\cite{DSAEurasip,maddpg_paper}. The latter structure has been shown to offer greater flexibility in modeling heterogeneous agents while leveraging a shared critic for more robust value estimation.

\paragraph{Neural Network Pruning}
With the growing adoption of IoT and edge computing, there is an increasing demand for running complex neural networks on memory-constrained embedded devices. To address this challenge, techniques such as neural network pruning \cite{toPruneorNotToPrune} have gained significant attention, as they reduce model size without substantial performance degradation. In the single-agent DRL domain, it has been well established that state-of-the-art algorithms often lead to agents underutilizing their parameters \cite{kumarUnderUtil}. 
Studies have shown that carefully designed dynamic pruning strategies can remove a significant portion of redundant weights/connections, while maintaining \cite{PoPsCohen}, or even improving \cite{DRL_sparse_good,stateSparseDRL}, the agent's performance. In~\cite{DRL_sparse_good}, \textit{gradual magnitude pruning} was presented as a general technique for improving performance in value-based DRL. However, these works primarily focus on single-agent environments where full or near-complete observability is assumed. 
%Studies have shown that applying standard pruning techniques can remove a significant portion of redundant weights/connections with minimal performance degradation \cite{stateSparseDRL,PoPsCohen}. Surprisingly, recent research has demonstrated that carefully designed dynamic pruning strategies can even enhance the performance of popular DRL algorithms \cite{toPruneorNotToPrune,DynamicSparseDRL,DRL_sparse_good}. Graesser et. al. \cite{stateSparseDRL} observe that applying \textit{gradual magnitude pruning} to a DQN \cite{DNQnature} agent, reduces the total number of parameters by 90\%, while improving achieved rewards by 50\%. Building on this previous study, Ceron et. al. \cite{DRL_sparse_good} suggest gradual magnitude pruning, as a general technique for improving performance in value-based DRL. However, these works primarily focus on single-agent environments where full or near-complete observability is assumed. 
Hence, pruning in multi-agent problems remains underexplored, especially for recurrent policies needed in partially observable environments.
% The added complexity of decentralized actor networks, heterogeneous agents, and time-correlated observations makes pruning significantly more challenging. 
A Pruning at Initialization (PaI) framework \cite{marlPrune} was introduced for MADRL, but it only prunes feed-forward agent networks, which are typically not appropriate for modeling temporal dependencies, as usually appearing in DSA problems \cite{DDSA-Cohen,DSAEurasip}.

\paragraph{Our contribution}\label{subsec:contribution}
Motivated by the recent findings of \cite{stateSparseDRL,DRL_sparse_good} we present, in this paper, a novel multi-agent actor pruning framework for distributed DSA. Each actor network is iteratively pruned using magnitude-based sparsification, guided by a pruning scheduler. Our numerical results showcase that our method not only preserves performance, but can even surpass dense IAGC-trained agents, reinforcing the role of pruning as a \textit{general performance-enhancing technique} in MADRL. In addition, we introduce a novel \textit{harmonic annealing} scheduler that enables periodic network weight regrowth to counteract excessive sparsification. This scheduler achieves comparable or superior performance at high sparsities compared to conventional linear and polynomial pruning schedules. Notably, it also has the potential to discover significantly better policies under favorable training conditions, highlighting the impact of dynamic sparsity in MARL training and policy optimization.

\section{System model}\label{sec:generalmodel}
Consider $N$ users sharing $K$ orthogonal channels (i.e., subbands). Each user (agent) $n$, at time $t$, can access one of the channels or even remain inactive (action $a_t^n$). The channel indices are $1,2,\ldots,K$ and an inactive action is represented with $a_t^n=0$. 
If a collision occurs (which means that $a_t^n=a_t^m$ for some $n,m$), neither of the users accesses that channel \cite{DDSA-Cohen}. Note that actions are taken in a fully decentralized/independent manner, and each user cannot know which channels the other users will select. Similarly to \cite{DDSA-Cohen}, we assume that all users  are constantly backlogged and can transmit on discrete time slots (i.e., Aloha-type narrowband transmissions).

The received signal after attempting channel access will be $o^n_t \in \{\text{ACK},\text{NACK}\}$, depending on whether a collision occurs. We will use $1$ to denote ACK, and $-1$ to denote NACK. If a user $n$ successfully accesses the $k$-th channel, transmission happens with an SNR of $\beta^n_k$. The SNR values are assumed to be time-invariant, but unknown \cite{DSAEurasip}. The cumulative SNR of user $n$ for a known and finite horizon $T$ is defined as:
\begin{equation}\label{eq:userSNR}
    \text{SNR}^n(T)\triangleq\sum_{t=1}^{T} \beta^n(t),
\end{equation}
where $\beta^n(t)\triangleq\beta^n_{a_t^n} \mathbbm{1}\{o^n_t=1\}$ with $\mathbbm{1}\{\cdot\}$ being the indicator function of a set/action. Note that the channels are not necessarily identical, since certain users may get higher SNR values on some channels depending on factors, such as the locations of the communication nodes and their equipment. In addition, the users are unaware of the access attempts of their peers, hence, they must learn their access patterns, while simultaneously identifying the most favorable channels based on their past observations and rewards.

Let $g^n$ represent the action selection policy of user $n$, and let ${\rm g}\triangleq(g^1,\ldots,g^N)$. In this work, we will try to solve the following \textit{throughput} maximization problem:
\begin{align*}
%\label{eq:throughput_objective}
 \mathcal{OP}: \quad  \max_{\rm g} E\bigg{[}\sum_{t=1}^T \sum_{n=1}^N \log_2\left(1+\beta^n(t)\right)\bigg{]}.
\end{align*}
%where $T$ is the time-horizon of the game.
The users in this system could be lightweight devices with limited memory, computational capabilities, and energy resources, such as IoT sensors, wearable devices, or smart home appliances. These constraints make it impractical for each user to employ large, or computationally intensive, models for optimizing their action selection policies. To address this, we leverage neural network pruning techniques, where unnecessary weights of an elaborate dense model are removed, obtaining a sparse model with considerably lower memory. 

% Besides the limited memory requirements of  the sparse neural network, sparse matrix multiplication operations required by its operation 

% complex and computationally expensive model (teacher model) is used to train smaller, efficient models (student models) suitable for deployment on resource-constrained devices. This ensures that the users can efficiently learn to optimize their channel access policies while maintaining a low computational footprint.

The model presented above can be easily extended to account for environments with Primary Users (PUs), leading naturally to the licensed shared access paradigm \cite{alexandDSAref1,alexandDSAref2}. Each PU can be actually modeled as an external process (e.g., \cite{zhaoPODMPGreedi,banditsDSA2,DDSA-Cohen}), and when a secondary user (agent) $n$ attempts to access an occupied channel $k$, it receives an observation indicating PU activity. We denote channel occupancy by a PU with an observation value equal to $-2$. A channel $k$ is occupied by a PU with probability $\pi_k$, in which case all access attempts will result in observations of $-2$. Otherwise, the previously defined observation and SNR rules apply.

%\color{red}
% The following are extensions for future work.
%\begin{enumerate}
%    \item{Individual service constraints}
%    \item{Multiple channel access by each user at each $t$}
%    \item{Joint channel access and power control}
%    \item{Adversarial interference}
%\end{enumerate}
% I think the second is the easiest.

\section{MARL with 
Harmonic Annealing Pruning}\label{sec:SRMAPPOfDSA}
In this section, we present our multi-agent algorithm comprising multiple training iterations; let their number be $I_{\rm T}$. Each $i$-th iteration ($i=1,\ldots,I_{\rm T}$) includes three steps. The first step involves sampling of trajectory data, whereas, in the second step, we use these trajectories to update the neural networks via Proximal Policy Optimization (PPO)~\cite{schulmanPPO}. Once the updates are completed, we prune the resulting actor networks during the final third step.

\subsection{Neural Network Structures}\label{subsec:neuralnets}
We consider $N$ individual recurrent actor networks $\theta^1,\theta^2,\ldots,\theta^N$. The input to both the actor and critic networks is the latest action-observation pair, which we represent as:
\begin{equation}
    \boldsymbol{y}^n_t\triangleq [a_{t-1}^n,o_{t-1}^n].
\end{equation}
All actors as well as the global critic $\phi$ are Long Short-Term Memory (LSTM) \cite{lstm} networks, whose final output is passed through a linear layer. For the actors, the linear layer's output passes through a softmax activation function to generate the action distributions. Each of these distributions is denoted as:
\begin{equation}
    g_t^{\theta^n}=g(a_t^n|\boldsymbol{y}^n_t,\boldsymbol{h}_{t-1}^n;\theta^n),
\end{equation}
with $g(\cdot)$ indicating each action's distribution dependence on the input parameters, and $\boldsymbol{h}_t^n$ being the recurrent network's hidden state vector. The global critic estimates the value for a given state. Therefore, its output
\begin{math}
    V(\boldsymbol{y}_{t}^n|\boldsymbol{h}_{t-1}^{n,c};\phi)
\end{math}
is a scalar, with $\boldsymbol{h}_{t}^{n,c}$ denoting its hidden state. 

% Our algorithm also uses target networks $\bar{\phi}^1,\bar{\phi}^2$ to enhance training stability. Finally, each agent will have its own replay buffer $\mathcal{B}^n$, that stores entire trajectories for training. One important feature of our algorithm, that differs from the feed-forward SAC for fully observed problems, is that the parameter updates are conducted by sampling data from entire, unified trajectories. Updates begin at the start of the episode, and carry on for its entire duration. This procedure is the \textit{bootstrapped sequential update} methodology first presented in \cite{stoneRDQN}.
%Before explaining how the networks are optimized, we will present our problem-specific artificial simulation. Trajectories from that simulation will be used to train the LSTM networks.

\subsection{Step 1: Trajectory Data Generation}\label{subsec:simulation}
We have designed a simulator generating episodes that describe the DSA system presented in Section~\ref{sec:generalmodel}. At the beginning of each episode, all agents receive an initial dummy observation: $o^n_0 = 0$ $\forall n$. The hidden states of both the actor and critic networks are initialized to zero. At each time step $t = 1,2,\dots,T$, all agents select actions $a_t^n$, and the simulator computes the resulting collisions among users, as well as the individual SNRs $\beta^n(t)$. The instantaneous reward provided to each $n$-th agent was the system throughput\footnote{Our algorithm is purely model-free, and hence, alternative rewards suitable for optimizing other objectives can be incorporated.}, given as follows:
\begin{equation}\label{eq:jointRew}
    r^n_t = \sum_{n=1}^N \log_2\left(1+\beta^n(t)\right).
\end{equation}
To increase training stability, we have performed reward normalization.

\subsection{Step 2: Neural Networks Updating}
By using multiple trajectories from the first step, we optimize the actors and the global critic with the clipped version of the PPO algorithm. We adopt the \textit{bootstrapped sequential updates} paradigm, introduced in \cite{stoneRDQN}, where updates are performed by initializing the hidden states at the beginning of each sampled trajectory, and sequentially updating the network with the current losses at each $t$, until the episode concludes. For a given trajectory and time instant $t$, the actor and critic gradient update rules (performed through the Adam optimizer \cite{adam}) are given by the following expressions:  
\begin{align}
   \theta^n &\gets \theta^n + \alpha_n \nabla_{\theta} L^{\text{clip}}(\boldsymbol{y}_t^n,a_t^n,\theta^n_{i-1},\theta^n),\\
    \phi &\gets \phi -\alpha_c \nabla_\phi L^n_{\rm critic}(\boldsymbol{y}_t^n|\boldsymbol{h}_{t-1}^{n,c};\phi),
\end{align}
where $\alpha_n$ and $\alpha_c$ represent learning rates, and the loses $L^{\rm clip}(\cdot)$ and $L_{\rm critic}^n(\cdot)$ were designed as follows. 

\textit{Actor Loss:}
Let us first define the clipping function as follows:
\begin{equation}
    \text{clip}(\epsilon,A)=\begin{cases}
        (1+\epsilon)A, & A \geq 0 \\
        (1-\epsilon)A, & A \leq 0
    \end{cases},
\end{equation}
with $\epsilon$ being a manually selected parameter.
The clip objective is then defined as 
\begin{align}
\label{eq:clip_objective}
    &L^{\text{clip}}(\boldsymbol{y}_t^n,a_t^n,\theta^n_{i-1},\theta^n)\triangleq\nonumber& \\
    &\min\bigg{(} 
    \frac{g(a_t^n|\boldsymbol{y}_t^n,\boldsymbol{h}_{t-1}^n;\theta^n)}{g(a_t^n|\boldsymbol{y}_t^n,\boldsymbol{h}_{t-1}^n;\theta^n_{i-1})} A^{g^{\theta^n_{i-1}}}(\boldsymbol{y}_t^n,a_t^n), \nonumber& \\
    & \hspace{3.25cm} \text{clip}(\epsilon,A^{g^{\theta^n_{i-1}}}(\boldsymbol{y}_t^n,a_t^n))
    \bigg{)}.
\end{align}
The advantage function $A^{g^{\theta^n_{i-1}}}(\boldsymbol{y}_t^n,a_t^n)$ under policy $g^{\theta^n_{i-1}}$ in~\eqref{eq:clip_objective} provides a way to compare the relative benefit of actions in a given state. Let $Q^{g^{\theta^n_{i-1}}}(\boldsymbol{y}_t^n,a_t^n)$ denote the action-value (often referred to as $Q$-value) function, which represents the expected cumulative reward after taking action $a_t^n$ at state $\boldsymbol{y}_t^n$, and then following $g^{\theta^n_{i-1}}$. The latter function is defined as:
\begin{equation}
    \label{eq:advDef}
    A^{g^{\theta^n_{i-1}}}(\boldsymbol{y}_t^n,a_t^n)  \triangleq Q^{g^{\theta^n_{i-1}}}(\boldsymbol{y}_t^n,a_t^n) - V^{g^{\theta^n_{i-1}}}(\boldsymbol{y}_t^n),
\end{equation}
which can be computed using \cite{GAEPaper}'s generalized estimator.

\textit{Critic Loss:}
For each $n$-th agent, the global critic's loss at each time $t$ is a typical least squares error function: \begin{equation}
    L_{\rm critic}^n(\mathbf{y}_t^n|\mathbf{h}_{t-1}^{n,c};\phi)\triangleq \left(V(\boldsymbol{y}_t^n|\mathbf{h}_{t-1}^{n,c};\phi)-\hat{R}_t^n \right)^2,
\end{equation}
where $\hat{R}_t^n $  denotes the reward-to-go at time $t$.

\subsection{Step 3: Neural Networks Pruning}
In this step, we apply \textit{magnitude-based unstructured pruning} to the actor networks of step $2$, by setting some of their weights to zero, achieving a sparsity level of $p_i^n$. This sparsity level can be determined by the gradual pruning rule \cite{toPruneorNotToPrune}:
%a pruning fraction \( p_i^n \), which follows the gradual pruning rule \cite{toPruneorNotToPrune}, according to which:
\begin{equation}
    p_i^n=\begin{cases}
        0 \quad \text{if } \quad i< i_{\rm{start}}\\
       p_{\rm final}^n\sigma^n(i)  \quad \text{if } \quad  i_{\rm{start}} \leq i <i_P,
    \end{cases},
\end{equation}
where $i_P$ denotes the total number of pruning iterations, $i_{\rm start}$ is the training iteration when pruning begins, $\sigma^n(\cdot)$ represents the pruning schedule, and $p_{\rm final}^n$ is the target sparsity level for $n$-th agent; clearly, we allow different sparsity requirements for each agent. Some agents may operate on lightweight, resource-constrained devices, necessitating smaller networks, whereas others with access to greater computational resources can accommodate larger models.

Two typical choices for scheduling sparsity levels are the linear and the third-degree polynomial schedulers~\cite{toPruneorNotToPrune,stateSparseDRL}:
\begin{equation}
\label{eq:baseline_schedulers}
\sigma_{\text{linear}}^n(i) \triangleq \frac{i-i_{\text{start}}}{i_P-i_{\text{start}}},\;
\sigma_{\text{poly}}^n(i)\triangleq 1-\left(1-\frac{i-i_{\text{start}}}{i_P-i_{\text{start}}}\right)^3.
\end{equation}
In these schedulers, sparsity monotonically increases at each $i$-th iteration. This implies that, once critical weights are pruned (i.e., set to zero), they cannot be recovered. This fact can potentially break crucial network pathways, thus, limiting the network's capacity for exploration, restricting its ability to adapt and refine policies over time. To deal with this limitation, in this paper, we propose the following novel, \textit{harmonic annealing} pruning scheduler, which selects sparsity levels as:
\begin{equation}
p_i^n=\min\Bigl\{\max\Bigl(b_i^n+c_i^n,\,p_0^n\Bigr),\,p_{\text{final}}^n\Bigr\},
\end{equation}
where
\begin{align}
c_i^n &= 0.1\,\sin\Bigl(\frac{2\pi\,i}{200}\Bigr),\\[1mm]
b_i^n &= p_{\text{final}}^n + \frac{1}{2}\Bigl(p_0^n-p_{\text{final}}^n\Bigr)\left(1+\cos\Bigl(\pi\,\frac{i - i_{\text{start}}}{i_P - i_{\text{start}}}\Bigr)\right).
\end{align}
In this definition, the parameter \(b_i^n\) follows a cosine annealing function that ensures smooth and gradual pruning, while \(c_i^n\) introduces a periodic oscillation to allow periodic weight regrowth during training. Finally, the parameter \( p_0^n \) represents the initial sparsity level of each $n$-th agent.

\begin{figure}
    \centering \includegraphics[width=0.9\linewidth]{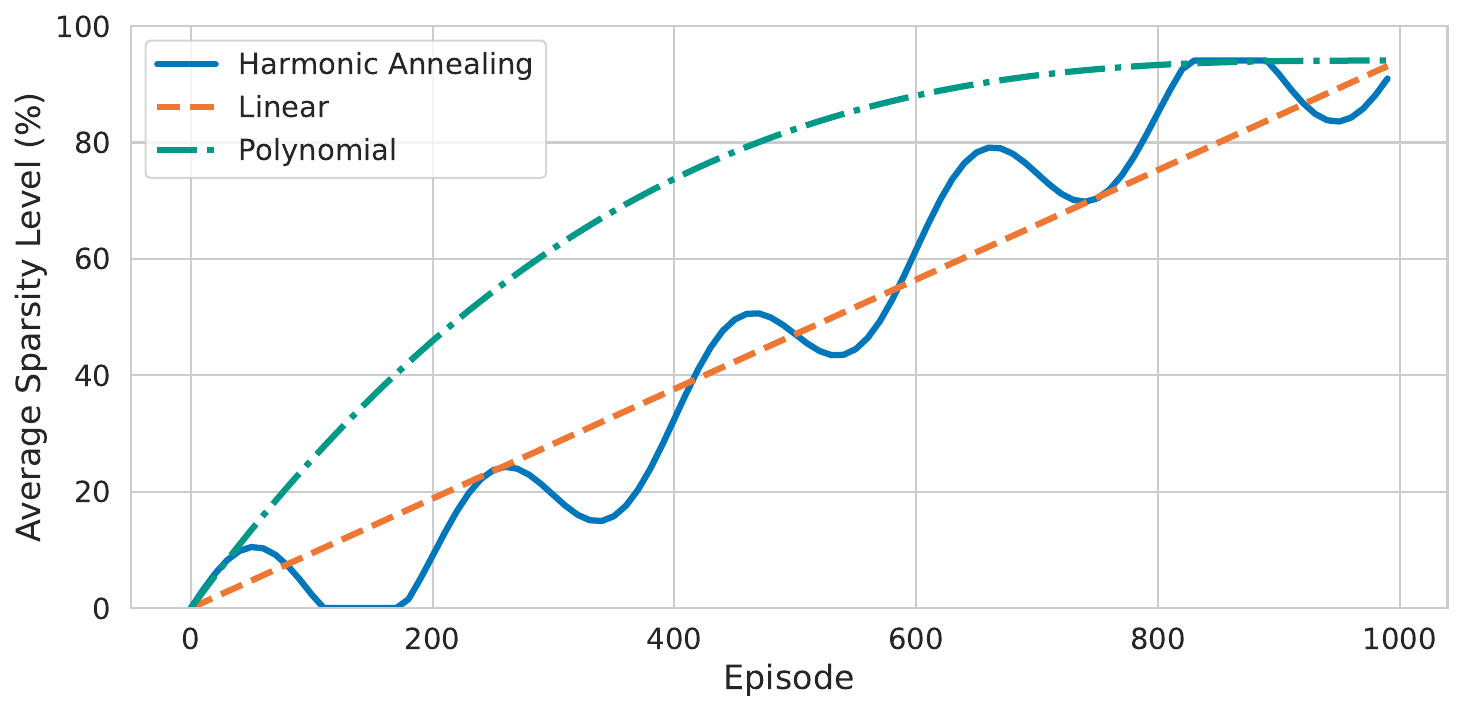}
    \caption{The average sparsity level for the three different considered pruning schedulers for $I_{\rm T}=1000$ and $p_{\rm final}=0.95$.}
    \label{fig:sparsity_schedulers}
\end{figure}

The average sparsity level across agents with the proposed harmonic annealing scheduler for $I_{\rm T}=1000$ and $p_{\rm final}=0.95$ is illustrated in Fig.~\ref{fig:sparsity_schedulers}, together with the schedulers in~\eqref{eq:baseline_schedulers}. It can be seen that our scheduler introduces periodic weight regrowth through its oscillatory nature, allowing previously pruned connections to recover. As it will be showcased in the following section, our dynamic prune-and-regrow mechanism, not only stabilizes training, acting as an implicit regularization mechanism, but also expands the search space, increasing the potential to uncover superior policies.

\paragraph*{Remark 1 (Pruning Critic Networks)} During deployment, only the actor networks are running at the agents, allowing the critic networks to be large, if this benefits the training process. Since the computational cost of large critics is incurred only during training, we restrict pruning to the actor networks. Moreover, in the single-agent setting, it has been observed that estimating value functions is more challenging than learning policies, and pruning critics often degrades performance~\cite{stateSparseDRL}.

\paragraph*{Remark 2 (Pruning Interval)}
Rather than pruning after each $i$-th iteration, we have noticed that allowing networks to regrow for a sufficient period, and then pruning every \(i_{\text{prune}}\) iterations can lead to a noticeable performance increase.

\section{Numerical Results and Discussion}
\subsection{Neural Networks Implementation Details}
We have used the PyTorch framework \cite{pytorch}. The actors were modeled as unidirectional LSTMs with two hidden layers of $128$ units, using Rectified Linear Units (ReLU) activations. The critic shared the same architecture, except for its final output layer. We have set the learning rates of the actors to \(\alpha_n=10^{-4}\) and those of the critic to \(\alpha_c=5 \times 10^{-5}\). The PPO's clip parameter was set to $\epsilon=0.2$ and the discount factor to $0.99$. The target sparsity levels of the agents were set as \(p_{\rm final}^n=0.95\) $\forall n=1,\ldots,N$. Each learning algorithm was run for $I_{\rm T}=1000$ episodes with a horizon of \(T=100\) time steps, and the results were obtained via averaging over $10$ different random seeds. Pruning was applied every \(i_{\text{prune}}=5\) iterations. For both the linear and harmonic annealing schedulers, we set \(i_{\rm start}=0\), while for the polynomial scheduler, we performed experiments with \(i_{\rm start} \in \{0, 200\}\). 
% Finally, for the harmonic annealing scheduler we set $c_a^n=1, c_l^n=200 \quad \forall n$.  
Further investigations on the pruning hyperparameters as well as sensitivity studies will be provided in the journal version of this work.

\subsection{Benchmark Schemes}
In terms of DSA benchmarks, we have considered two learning-based ones: a recurrent DQN  with full parameter sharing between the agents \cite{DDSA-Cohen}, and a recurrent actor-critic with individual actors and a global critic (IAGC), as in \cite{DSAEurasip,maddpg_paper}, as well as a randomized slotted aloha scheme. For fair comparison, all methods used identical network architectures.

In addition, to evaluate the effectiveness of our proposed pruning framework, we compare it with the \textit{PaI with Parameter Sharing} baseline~\cite{marlPrune}. In this approach, all actor networks were initialized with identical parameters, but each agent was assigned a unique pruning mask. The agents were then trained using PPO, with only the remaining parameters being shared. The target sparsity was set to $50\%$, which is lower than that of our proposed method, since we observed that, at higher sparsity levels, PaI's performance deteriorated significantly, approaching that of a random policy.

\subsection{Performance Evaluation}
We have simulated a DSA system comprising $N=10$ secondary users (agents) and $K=N/2$ orthogonal channels. Studies with larger $N$ values will be presented in a future journal version of this work. The SNR values were sampled as $\beta_k^n \sim \mathcal{U}[30,40]$ at the start of each episode, with a communication horizon of $T=100$. We have considered two different system setups: (A) without PUs; and (B) each channel $k$ was occupied by a PU with probability $\pi_k=0.2$, independently of the other channels.

\begin{figure}[!t]
    \centering
    \begin{subfigure}{.45\textwidth}
        \hspace{-2.2em}
        \centering
        \includegraphics[width=\textwidth]{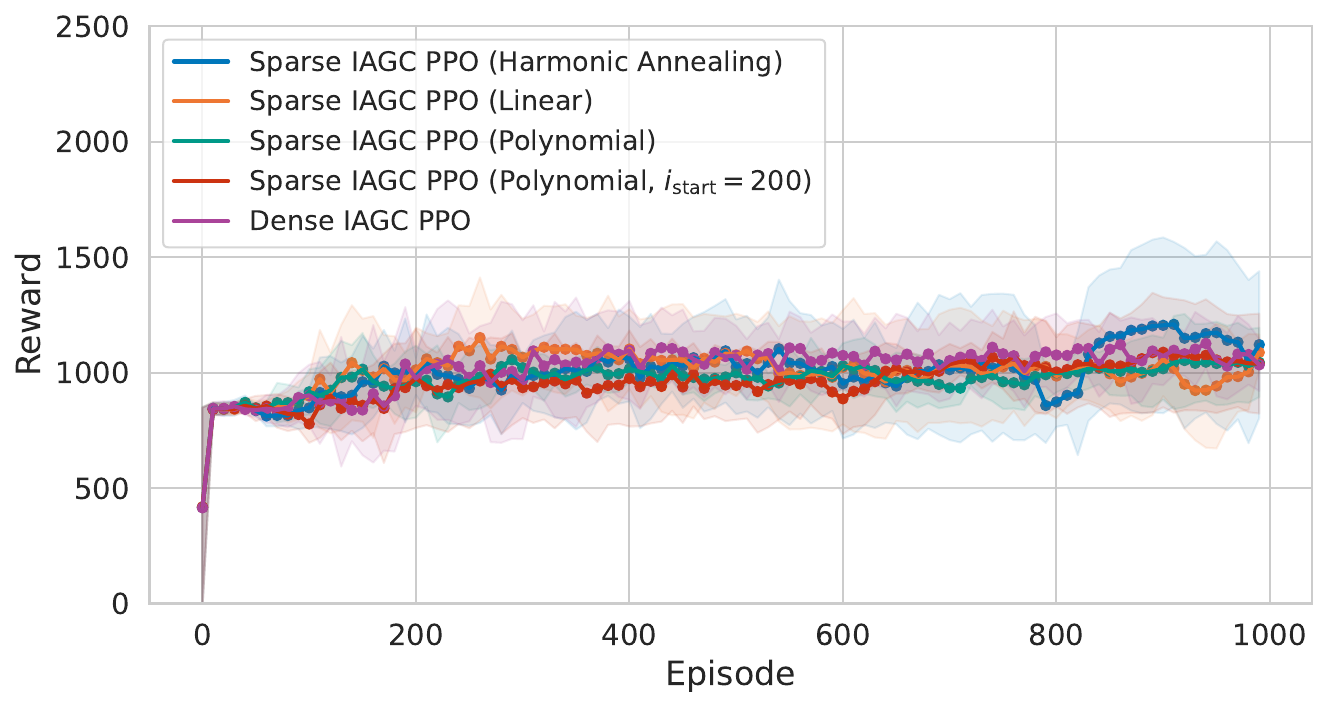}
        \caption{Setup A.}
        \vspace{0.3em}
    \end{subfigure}
    \hfill
    \begin{subfigure}{.45\textwidth}
        \hspace{-2.2em}
        \centering
        \includegraphics[width=\textwidth]{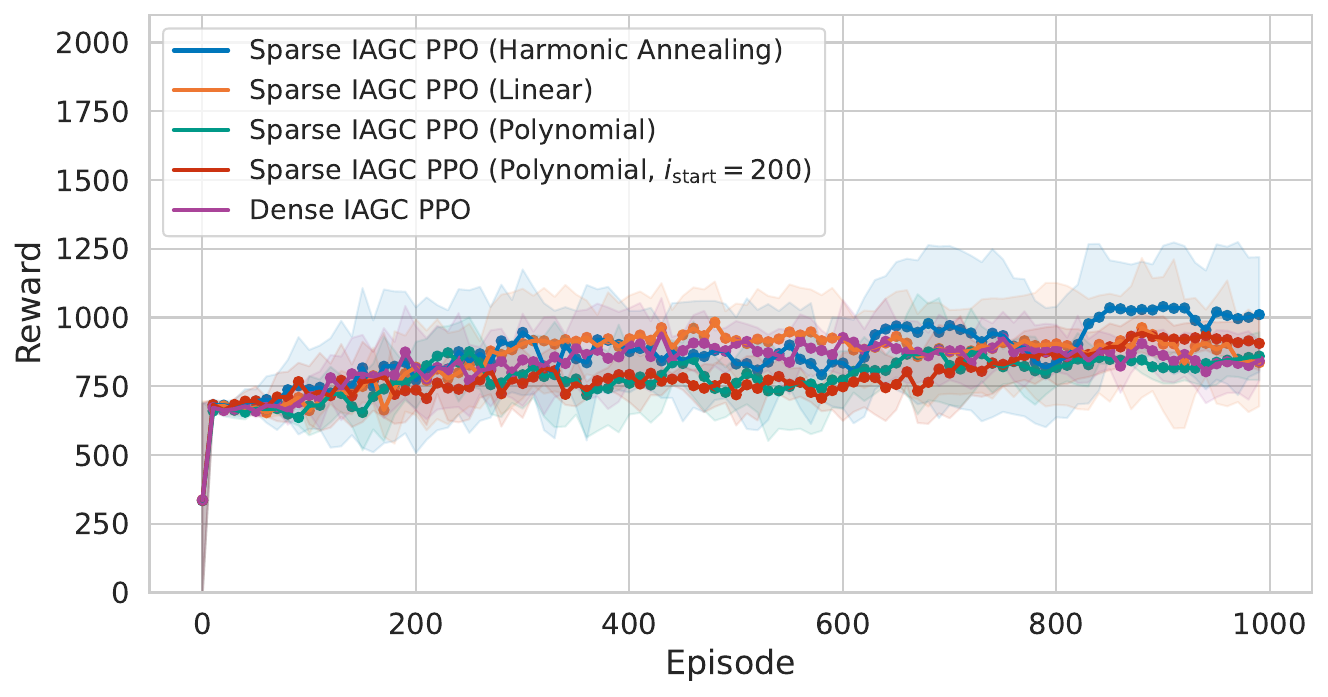}
        \caption{Setup B.}
    \end{subfigure}
    \caption{Training curves for the proposed MADRL framework for all considered pruning schedulers, including the standard deviation across multiple random seeds (shaded region).}
    \label{fig:results}
\end{figure}

At every \(10\) episodes, the networks training was paused for evaluation over \(100\) episodes. Figures \ref{fig:sparsity_schedulers}  and \ref{fig:results} include respectively the sparsity levels and training curves of all considered schedulers, illustrating both the mean episodic reward ($\frac{1}{N}\sum_{t=1}^T \sum_{n=1}^N r_t^n $) and the reward's standard deviation across multiple seeds. Additionally, Table~\ref{tab:topperforming_results} reports the final rewards achieved by the best performing seed for each studied method. The key takeaways from our results are as follows:
% \begin{itemize}  
%     \item \textbf{Comparison with DSA benchmarks.}  
%     On average, both our sparse algorithms and the dense IAGC-PPO outperform the considered DSA benchmarks.  
%     In the first environment, all methods achieve an average reward exceeding \(1000\), while in the second environment, they surpass \(800\) after training convergence.  

%     \item \textbf{Comparison with pruning baseline.}  
%     All three variants of our algorithm outperform PaI despite achieving higher sparsities.  

%     \item \textbf{Evaluation of pruning schedulers.}  
%     All pruning schedules achieve \textit{highly sparse networks} (\(\geq 90\%\) sparsity) with minimal performance degradation.  
%     However, our harmonic scheduler exhibits superior average performance at high sparsities and is particularly effective at \textit{discovering the best policies under favorable seeds}.  
% \end{itemize}

\textbf{Comparison with DSA benchmarks:} Both dense (i.e., unpruned) IAGC-PPO and all our sparse IAGC PPO ($95\%$) versions with different pruning scheduling outperform the considered DSA benchmarks. In setup (A), all methods achieve an average reward exceeding \(1000\), while in Setup (B), all of surpass \(800\) after training convergence.  

\textbf{Comparison with PaI pruning baseline:} It can be observed that all three variants of our algorithm outperform PaI in terms of reward, despite achieving higher sparsities.  

\textbf{Comparison among pruning schedulers:} As shown, all pruning schedules achieve \textit{highly sparse networks} (i.e., \(\geq 90\%\) sparsity) with minimal performance degradation. Notably, our harmonic scheduler exhibits superior average performance at high sparsities, and is particularly effective at \textit{discovering the best policies under favorable seeds}.

% \begin{figure}[!t]
%     \begin{subfigure}{.35\textwidth}
%         \scalebox{1.2}{\includegraphics[width=\textwidth]{figures/test_reward_with_variability_default_env.pdf}}
%         \caption{Environment A}
%     \end{subfigure} 
%     \begin{subfigure}{.35\textwidth}
%         \scalebox{1.2}{ \includegraphics[width=\textwidth]{figures/test_reward_with_variability_random_env.pdf}}
%         \caption{Environment B}
%     \end{subfigure}%
%     \caption{ Average rewards, and variances computed over multiple seeds.}\label{fig:results}
% \end{figure}

\begin{table}[h]
\centering
\setlength{\tabcolsep}{2pt} % Adjust column spacing
\renewcommand{\arraystretch}{1.3} % Increase row spacing
\caption{Reward evaluation for the best-case seed.}
\begin{tabular}{lcc}
\toprule
\textbf{Algorithms} & \textbf{Setup A} & \textbf{Setup B} \\
\midrule
Slotted Aloha & 993.77 & 764.10 \\
IAGC \cite{DSAEurasip,maddpg_paper} & 996.11 & 802.24 \\
DQN with parameter sharing \cite{DDSA-Cohen} & 742.54 & 555.41 \\
\midrule
PaI \cite{marlPrune} (50\%) & 1111.88 & 825.34 \\
Dense IAGC PPO & 1556.83 & 1115.74 \\
\midrule
\multicolumn{3}{l}{Proposed IAGC PPO (95\%) with scheduler: } \\
\quad Harmonic annealing  & \textbf{1826.86} & \textbf{1223.41} \\
\quad Linear & 1194.82 & 1083.76 \\
\quad Polynomial & 1412.89 & 1092.19 \\
\quad Polynomial ($i_{\rm start}=200$) & 1496.46 & 1048.51 \\
\bottomrule
\end{tabular}
\label{tab:topperforming_results}
\end{table}

\section{Conclusion and future work}
In this paper, we presented a novel multi-agent gradual pruning approach based on the IAGC framework and applied it to the distributed DSA problem. In addition to existing pruning schedulers, we proposed a harmonic annealing scheduler that enables periodic weight regrowth. Our experiments demonstrated that our framework outperforms existing MADRL pruning schemes as well as DSA benchmarks. Notably, harmonic pruning not only achieves strong performance, but also unveils optimal policies missed by other pruning schedulers. Future work includes exploring alternative objective functions and extensions of our approach to other applications, such as games~\cite{Hanabi} and transmit power control \cite{DDSAandControl}. Additionally, we plan to investigate meta-learning techniques to bias training towards promising initialization conditions.

\bibliographystyle{ieeetr}
\bibliography{references}
\end{document}